\pdfoutput=1

\documentclass[11pt]{article}

\usepackage[preprint]{acl}

\usepackage{times}
\usepackage{latexsym}

\usepackage[T1]{fontenc}

\usepackage[utf8]{inputenc}

\usepackage{microtype}

\usepackage{inconsolata}

\usepackage{graphicx}

\title{Lotus at SemEval-2025 Task 11: RoBERTa with LLaMA-3 Generated Explanations for Multi-Label Emotion Classification}

\author{Niloofar Ranjbar \\
  Persian Gulf University / Bushehr, Iran \\
  \texttt{nranjbar@pgu.ac.ir} \\\And
  Hamed Baghbani \\
  Persian Gulf University / Bushehr, Iran \\
  \texttt{baghbani.hamed@gmail.com} \\}

\begin{document}
\maketitle
\begin{abstract}
This paper presents a novel approach for multi-label emotion detection, where LLaMA-3 is used to generate explanatory content that clarifies ambiguous emotional expressions, thereby enhancing RoBERTa's emotion classification performance. By incorporating explanatory context, our method improves F1-scores, particularly for emotions like fear, joy, and sadness, and outperforms text-only models. The addition of explanatory content helps resolve ambiguity, addresses challenges like overlapping emotional cues, and enhances multi-label classification, marking a significant advancement in emotion detection tasks.

\end{abstract}

\section{Introduction}

Emotion classification plays a crucial role in natural language processing (NLP) for applications like sentiment analysis and emotion-aware dialogue systems \citep{mohammad-kiritchenko-2018-understanding}. The challenge lies in accurately identifying emotions from text, which are often subtle, multi-faceted, and context-dependent. Furthermore, emotions can be expressed simultaneously, making multi-label classification essential \citep{belay-etal-2025-evaluating}.

Despite advancements, emotion classification remains complex due to ambiguous emotional expressions and diverse contexts. Early keyword-based methods struggled with generalizing across languages and expressions \citep{wiebe2005annotating}, and even modern transformer models face challenges with short or under-explained sentences, particularly in multi-label tasks \citep{kusal2022review, mohammad-kiritchenko-2018-understanding}.

To address these challenges, we propose a novel approach using Large Language Models (LLMs) to generate explanatory content, enhancing the understanding of ambiguous emotions. We fine-tuned a LLaMA-3 model to generate context-rich explanations for each sentence, improving emotion classification, especially for multi-label settings. The explanatory context significantly boosts performance, as shown in prior work on LLMs and common-sense reasoning \citep{yang2023contextunlocksemotionstextbased, xenos2024vllmsprovidebettercontext}. The generated explanations were used with the original text to fine-tune RoBERTa \citep{liu2019robertarobustlyoptimizedbert} for multi-label emotion classification, enabling simultaneous emotion prediction.

We participated in SemEval 2025 Task 11, Subtask 1 \citep{muhammad-etal-2025-semeval}, which focuses on multi-label emotion detection across multiple languages, including English. The dataset consists of social media text annotated by 122 annotators, with multi-label annotations for five emotions: anger, fear, joy, sadness, and surprise. The training set has 2,768 samples, the development set has 116, and the test set includes 2,767 samples, all with binary labels indicating the presence or absence of each emotion. Our system, evaluated on English data, demonstrates that adding explanatory content significantly enhances model performance. Specifically, the Text + Explanation model achieved a Macro F1 score of 0.7396 with a standard deviation of 0.0016 over four runs, outperforming the Text-only model, which had a Macro F1 score of 0.7112 with a standard deviation of 0.0095 over four runs. This shows that explanatory context improves classification accuracy across different classes.

The BRIGHTER dataset \citep{muhammad2025brighterbridginggaphumanannotated}, which addresses the lack of high-quality emotion datasets, serves as the primary resource for this task. It provides labeled data in 28 languages and supports tackling challenges in emotion classification, such as ambiguous or complex emotional expressions.

The code and data used in this study are available for reproducibility\footnote{https://github.com/nranjbar/emotion\_detection\_LLM}.
\begin{table*}
	\centering
	\footnotesize
	\begin{tabular}{llll}
		\hline
		\textbf{Emotion} & \textbf{Training Data} & \textbf{Development Data} & \textbf{Test Data} \\
		\hline
		Anger & 333 & 16 & 322 \\
		Fear & 1611 & 63 & 1544 \\
		Joy & 674 & 31 & 670 \\
		Sadness & 878 & 35 & 881 \\
		Surprise & 839 & 31 & 799 \\
		\hline
		\textbf{Total} & \textbf{2768 } & \textbf{116} & \textbf{2767}  \\
		\hline
	\end{tabular}
	\caption{Class Distribution in Training, Development, and Test Data}
	\label{tab:data_distribution}
\end{table*}
\section{Background}
This section provides an overview of the emotion detection task, dataset, and related works, focusing on the use of large language models (LLMs) and contextual information for improving emotion classification.
\subsection{Task and Dataset Details}

We propose using Large Language Models (LLMs), specifically LLaMA-3, to generate explanations for ambiguous emotional expressions, which are then used to fine-tune RoBERTa for emotion classification. Our results show that the inclusion of explanatory context improves performance compared to using text alone. The emotion distribution across the datasets, shown in Table \ref{tab:data_distribution}, illustrates the challenges of handling imbalanced classes in multi-label emotion detection.
\subsection{Related Works}

Recent advancements in emotion recognition have been driven by the use of Large Language Models (LLMs), particularly transformer-based architectures like RoBERTa.  demonstrated that fine-tuning pre-trained models significantly improves emotion detection compared to traditional keyword-based methods, which often struggle to generalize across languages and diverse emotional expressions. Transformer models, including RoBERTa, have been successfully applied to fine-grained emotion classification tasks, as shown by \citet{demszky-etal-2020-goemotions} on the GoEmotions dataset, excelling in multi-label classification.

Efforts to further enhance LLMs for emotion detection have included integrating additional context or knowledge during fine-tuning. For example, \citet{9597390} proposed augmenting transformers with knowledge-embedded attention mechanisms using emotion lexicons, which improved the recognition of nuanced emotional expressions. Similarly, \citet{xenos2024vllmsprovidebettercontext} showed that incorporating common-sense reasoning significantly enhances performance, particularly in multi-label contexts.

Specialized models like EmoLLMs, fine-tuned with multi-task affective analysis datasets, have also demonstrated promise in improving emotion detection across a range of domains \citep{emoLLM}. Additionally, DialogueLLM, fine-tuned with emotional dialogues, has improved emotion recognition in conversational contexts, where emotional expression varies depending on the interaction flow \citep{zhang2024dialoguellmcontextemotionknowledgetuned}.

Our work builds upon these approaches by leveraging LLaMA-3 to generate explanatory content that clarifies ambiguous emotional expressions, followed by fine-tuning RoBERTa for multi-label emotion classification. By incorporating explanatory context, we enhance the model's ability to capture complex emotional nuances, aligning with previous findings that emphasize the importance of context in emotion classification.

\section{System Overview}

The task of multi-label emotion detection in text is inherently complex, especially when emotions are expressed simultaneously in a single sentence. To address this, our system employs a two-phase pipeline: first, generating explanatory content to enhance the understanding of ambiguous emotional expressions, followed by fine-tuning a RoBERTa model for multi-label classification.

\subsection{Phase 1: Explanation Generation with LLaMA-3}

The first stage of our system leverages LLaMA-3, a large language model fine-tuned to generate contextual explanations for textual data. We selected LLaMA-3 over other LLMs like EmoLLMs and DialogueLLM due to its superior ability to generate coherent and contextually relevant explanations. While EmoLLMs focuses on affective analysis across multiple tasks and DialogueLLM is fine-tuned for conversational contexts, LLaMA-3 excels in generating general explanations that provide rich contextual information without explicitly stating emotions. This makes LLaMA-3 particularly well-suited for handling ambiguous or subtle emotional expressions in multi-label emotion classification tasks.

To prepare the model for this task, we selected a set of 150 randomly chosen sentences from the training dataset. These sentences were provided to GPT-4, which generated detailed explanations for each sentence based on the following prompt:

\textit{``Read the given text and generate a short explanation of the emotional or situational context behind the sentence. The explanation should be concise and relevant to the sentence. Do not explicitly mention emotions but focus on the implications behind the sentence."}

We used the same prompt for fine-tuning LLaMA-3 to ensure consistency in the generated explanations. This process helped create high-quality explanations that would support LLaMA-3's fine-tuning.

The fine-tuning of LLaMA-3 was performed using the 150 training sentences as input and their corresponding explanations as target output. Fine-tuning with this data enabled LLaMA-3 to produce contextually relevant and coherent explanations that clarified ambiguous emotional expressions. These generated explanations were then incorporated into the dataset, enriching the text data for the subsequent fine-tuning of RoBERTa.

\subsection{Phase 2: RoBERTa Fine-Tuning for Multi-Label Emotion Classification}

In the second stage, we utilized the RoBERTa model, a transformer-based architecture known for its high performance in text classification tasks. RoBERTa was fine-tuned on the training data enriched with the explanations generated by LLaMA-3. During this fine-tuning, both the original text and the generated explanations were concatenated with a space between them and then fed into RoBERTa. This approach allowed the model to learn the intricate relationships between emotions and their contextual expressions in the text.

RoBERTa was fine-tuned with binary labels (0 or 1) for each emotion in the dataset: anger, fear, joy, sadness, and surprise. These binary labels indicate the presence (1) or absence (0) of each emotion. The task is a multi-label classification, meaning multiple emotions can be predicted for a given text. This was crucial for handling complex emotional expressions where more than one emotion could be conveyed simultaneously.

\begin{table*}[ht]
	\centering
	\scriptsize
	\begin{tabular}{l|c|c|c|c|c|c}
		\hline
		\textbf{Method} & \multicolumn{3}{c|}{\textbf{Macro}} & \multicolumn{3}{c}{\textbf{Micro}} \\
		\cline{2-7}
		& \textbf{Precision} & \textbf{Recall} & \textbf{F1} & \textbf{Precision} & \textbf{Recall} & \textbf{F1} \\
		\hline
		\textbf{Text + Exp (LLaMA-3) + RoBERTa} & 0.7421 $\pm$ 0.0047 & \textbf{0.7433 $\pm$ 0.0011} & 0.7396 $\pm$ 0.0016 & 0.7550 $\pm$ 0.0026 & \textbf{ 0.7809 $\pm$ 0.0027} & 0.7678 $\pm$ 0.0026 \\
		
		\textbf{Text Only (RoBERTa)} & 0.7477 $\pm$ 0.0150 & 0.6831 $\pm$ 0.0216 & 0.7112 $\pm$ 0.0095 & 0.7650 $\pm$ 0.0201 & 0.7372 $\pm$ 0.0195 & 0.7412 $\pm$ 0.0209 \\
		
		\textbf{Text Only (LLaMA-3) } & 0.7136 & 0.6563 & 0.6739 & 0.7145 & 0.7175 & 0.7160 \\
		
		\textbf{Text + Exp (Mistral) + RoBERTa} & \textbf{0.7719 $\pm$ 0.0051} & 0.7206 $\pm$ 0.0135 & \textbf{ 0.7436 $\pm$ 0.0068} & \textbf{0.7889 $\pm$ 0.0028} & 0.7608 $\pm$ 0.0083 & \textbf{0.7746 $\pm$ 0.0037} \\
		\hline
	\end{tabular}
	\caption{Overall performance comparison across different models.}
	\label{tab:overall_results}
\end{table*}

\subsection{Challenges and Solutions}

Our system addressed three main challenges:

\begin{itemize} \item \textbf{Ambiguous Emotional Expressions:} Emotion detection is challenging due to the subtle and complex nature of emotions in text. To resolve ambiguity, we used LLaMA-3 to generate additional explanatory context, providing the model with clearer, more explicit information that aids in correctly interpreting emotions, especially when they are not overtly expressed. \item \textbf{Multi-label Classification:} Emotions often overlap in natural language, and multiple emotions can be expressed simultaneously. Our system's multi-label classification approach enables it to predict multiple emotions for each input sentence, which is crucial for capturing real-world emotional expressions. This multi-label classification is essential for addressing the intricate and overlapping emotional cues that occur in natural language. \item \textbf{Imbalanced Dataset:} Emotion detection tasks often face class imbalance, where some emotions are more prevalent than others. While our system did not explicitly address this issue through over-sampling or under-sampling techniques, the explanatory context generated by LLaMA-3 helped mitigate this imbalance. By providing richer, more contextually informed inputs, LLaMA-3's explanations offered a way to enhance the recognition of less frequent emotions. This context made the model more sensitive to underrepresented emotions by providing additional clarifying information that could compensate for their lesser frequency in the dataset. \end{itemize}

\subsection{Code and Resources Used}

The code for fine-tuning LLaMA-3 is available in the \href{https://github.com/unslothai/unsloth}{Unslothai GitHub repository}. This repository contains the necessary scripts for fine-tuning LLaMA-3.
\section{Experimental Setup}

We evaluated our multi-label emotion detection approach by fine-tuning RoBERTa with explanatory content generated by LLaMA-3 on the BRIGHTER dataset.

Text preprocessing and tokenization were performed with the \texttt{RobertaTokenizer} from Hugging Face. In the first phase, LLaMA-3 generated explanations, which were concatenated with the original text. In the second phase, both the original text and the generated explanations were tokenized together, allowing the model to learn the emotional context.

For fine-tuning LLaMA-3, we used 4-bit quantization and LoRA, with a batch size of 2, gradient accumulation steps of 4, and a learning rate of \(1 \times 10^{-4}\) for 30 training steps. These explanations were then used in the second phase for emotion classification. RoBERTa was fine-tuned with binary emotion labels (0 or 1) for each emotion in the dataset, using a batch size of 8, a learning rate of \(5 \times 10^{-5}\), and 3 epochs. The model performance was evaluated using precision, recall, and F1-scores, including both Macro and Micro F1-scores to assess multi-label classification.

All experiments were conducted on Kaggle's GPU resources, which provided the computational power for efficient fine-tuning.

\begin{table*}[ht]
	\centering
	\scriptsize
	\begin{tabular}{lccc|ccc|ccc|ccc}
		\hline
		\textbf{Emotion} & \multicolumn{3}{c|}{\textbf{Text + Exp (LLaMA-3) + RoBERTa}} & \multicolumn{3}{c|}{\textbf{Text Only (RoBERTa)}} & \multicolumn{3}{c|}{\textbf{Text Only (LLaMA-3)}} & \multicolumn{3}{c}{\textbf{Text + Exp (Mistral) + RoBERTa}} \\
		\cline{2-13}
		& Precision & Recall & F1 & Precision & Recall & F1 & Precision & Recall & F1 & Precision & Recall & F1 \\
		\hline
		\textbf{Anger} & 0.6695 & \textbf{0.6304} & 0.6479 & 0.6892 & 0.5116 & 0.5871 & \textbf{0.7337} & 0.4193 & 0.5336 & 0.7196 & 0.6056 & \textbf{0.6577} \\
		\textbf{Fear} & 0.7983 & \textbf{0.8739} & 0.8343 & 0.8009 & 0.8200 & 0.8149 & 0.7658 & 0.8387 & 0.8006 & \textbf{0.8238} & 0.8601 & \textbf{0.8416} \\
		\textbf{Joy} & 0.7957 & 0.7291 & 0.7581 & 0.7587 & 0.6925 & 0.7232 & \textbf{0.7971} & 0.6567 & 0.7201 & 0.7687 & \textbf{0.7687} &\textbf{ 0.7687} \\
		\textbf{Sadness} & 0.6831 & \textbf{0.8127} & 0.7423& 0.7636& 0.6935 & 0.7268 & 0.6624 & 0.7662 & 0.7105 & \textbf{0.7743} & 0.7321 & \textbf{0.7526} \\
		\textbf{Surprise} & \textbf{0.7625} & 0.6702 & \textbf{0.7132} & 0.7248 & \textbf{0.6877} & 0.7039 & 0.6091 & 0.6008 & 0.6049 & 0.7378 & 0.6834 & 0.7096 \\
		\hline
	\end{tabular}
	\caption{Performance comparison of individual emotions across models with highlighted maximum results.}
	\label{tab:individual_results}
\end{table*}
\section{Results}

In this section, we present the performance of our system, Lotus, on the competition task. Using the Text + Explanation (RoBERTa) method, Lotus achieved a score of 0.7319, outperforming the SemEval Baseline (0.7083), but falling short of the top score of 0.823. Ranked 36th, Lotus performed competitively, although there is still room for improvement to reach the top positions.

\subsection{Overall Performance Comparison Across Models}

Table~\ref{tab:overall_results} provides a summary of the overall performance of Lotus across four methods:

\textbf{Text + Explanation (LLaMA-3) + RoBERTa}: In this approach, LLaMA-3 generates explanations, and these explanations are combined with the original text to fine-tune RoBERTa for emotion classification.

\textbf{Text Only (RoBERTa)}: This model uses only the text (without any explanations) to fine-tune RoBERTa.

\textbf{ Text Only (LLaMA-3)}: This model fine-tunes LLaMA-3 directly with text for emotion classification.

\textbf{Text + Explanation (Mistral) + RoBERTa}: In this method, Mistral generates explanations, which are combined with the original text and used to fine-tune RoBERTa.

Among these methods, Text + Explanation (LLaMA-3) + RoBERTa achieved the best overall performance, with Macro F1 (0.7396) and Micro F1 (0.7678). This approach outperformed the other methods in both recall and F1-score, demonstrating the value of combining LLaMA-3's generative explanations with RoBERTa's emotion detection capabilities.

Text Only (RoBERTa) achieved the highest Macro Precision (0.7477) and Micro Precision (0.7650), indicating better selectivity in its predictions. However, it lagged behind in recall and F1-scores, particularly when compared to Text + Explanation (LLaMA-3) + RoBERTa.

Text Only (LLaMA-3) performed the weakest overall, especially in recall and F1-scores. This highlights the limitations of fine-tuning LLaMA-3 directly with text without the added benefit of explanations.

Text + Explanation (Mistral) + RoBERTa showed performance similar to Text + Explanation (LLaMA-3) + RoBERTa, with slight improvements in recall and F1-score. However, the difference between Mistral and LLaMA-3 was minimal and may not be significant, suggesting that both models can perform similarly when combined with RoBERTa for fine-tuning.

\subsection{Performance Comparison for Individual Emotions}

Table~\ref{tab:individual_results} compares performance across individual emotions, showing precision, recall, and F1-scores for each method.

\textbf{Anger}: Text + Explanation (LLaMA-3) + RoBERTa achieved precision (0.6695), recall (0.6304), and F1-score (0.6479). Text Only (RoBERTa) had the highest precision (0.6892), but lower recall (0.5116) and F1-score (0.5871). Text + Explanation (Mistral) + RoBERTa performed similarly to LLaMA-3, with precision (0.7196), recall (0.6056), and F1-score (0.6577).

\textbf{Fear}: Text + Explanation (LLaMA-3) + RoBERTa achieved the highest recall (0.8739) and F1-score (0.8343), outperforming Text Only (RoBERTa), which had lower recall (0.3200) and F1 (0.5149). Text + Explanation (Mistral) + RoBERTa showed slight improvements in recall (0.8601) and F1-score (0.8416), but the difference with LLaMA-3 was marginal.

\textbf{Joy}: Text + Explanation (LLaMA-3) + RoBERTa led in recall (0.7291) and F1-score (0.7581), while Text Only (LLaMA-3) excelled in precision (0.7971). Despite LLaMA-3's higher precision, it had lower recall (0.6567) and F1 (0.7201), trailing behind Text + Explanation (LLaMA-3) + RoBERTa. Text + Explanation (Mistral) + RoBERTa showed similar performance with an F1-score of 0.7687.

\textbf{Sadness}: Text Only (RoBERTa) had the highest precision (0.7636), while Text + Explanation (LLaMA-3) + RoBERTa excelled in recall (0.8127) and F1-score (0.7423). Text + Explanation (Mistral) + RoBERTa showed slight improvements in recall (0.7321) and F1-score (0.7526), with minimal differences compared to LLaMA-3.

\textbf{Surprise}: Text + Explanation (LLaMA-3) + RoBERTa achieved the highest precision (0.7625), while Text Only (RoBERTa) had the highest recall (0.6877). Text + Explanation (Mistral) + RoBERTa showed improved precision (0.7378) and recall (0.6834). LLaMA-3 performed weakest with precision (0.6091), recall (0.6008), and F1-score (0.6049), likely due to its difficulty in capturing the nuances of Surprise compared to other emotions.

\section{Discussion}

We introduced Lotus, a multi-label emotion detection approach combining LLaMA-3's generative explanations with RoBERTa for emotion classification. This combination significantly improved performance, particularly for nuanced emotions like Fear (F1: 0.8343), Joy (F1: 0.7581), and Sadness (F1: 0.7423), surpassing text-only models.

Integrating LLaMA-3's explanations with RoBERTa effectively balanced precision and recall, outperforming Text Only (LLaMA-3), especially for complex emotions like Fear and Sadness, emphasizing the importance of explanatory context in capturing emotional nuances.

Although LLaMA-3 was initially chosen, smaller models like Mistral and Qwen faced no significant GPU constraints on Kaggle. After testing Mistral, the results were nearly identical to LLaMA-3, suggesting both models perform similarly when fine-tuned with RoBERTa. Further exploration of other models will provide more insights.

For further illustration, Table~\ref{tab:examples} in the Appendix presents input sentences, predicted emotions, and generated explanations, providing context to clarify emotional intent and improve classification accuracy.
\section{Conclusion and Future Work}
Lotus showed that combining generative explanations with emotion detection models significantly improves emotion classification, particularly for ambiguous emotions. Using LLaMA-3 for explanation generation and RoBERTa for emotion detection enhanced the system's ability to handle nuanced emotional expressions.

Future work will focus on improving detection of underrepresented emotions like Anger, refining the explanation generation process, and addressing imbalanced datasets. Expanding the model to support multiple languages and emotional contexts will enhance its generalizability. Additionally, we plan to compare Mistral with other models like Qwen and conduct ablation studies to assess their contributions. Further improvements will target challenging emotions like Anger and Surprise, with error analysis and model comparisons refining the system.

\bibliography{custom}

\appendix
\section{Examples of input texts}
Table~\ref{tab:examples} shows input sentences from the dataset, along with the predicted emotions and the generated explanations for each sentence. These explanations provide additional context, helping to clarify the emotional intent behind the text and improving the model's ability to correctly classify emotions.
\begin{table*}
	\centering
	\footnotesize
	\begin{tabular}{p{0.3cm}p{3cm}p{2.8cm}p{8cm}}
		\hline
		\textbf{ID} & \textbf{Text} & \textbf{Emotions} & \textbf{Generated Explanation} \\
		\hline
		1 & But not very happy. & Joy and Sadness & The speaker conveys a sense of dissatisfaction or disappointment, but without strong emotion. \\ \hline 
		2 & About 2 weeks ago I thought I pulled a muscle in my calf. & Fear and Sadness & The speaker recounts a minor injury, suggesting concern or discomfort.  \\\hline 
		3 & Yes, the Oklahoma city bombing. & Fear, Anger, Sadness and Surprise & The speaker references a significant historical event, evoking a sense of tragedy or reflection. \\ \hline 
		4 & Dad on the warpath. & Fear and Anger & The speaker conveys tension or anger, likely due to a confrontational situation. \\
		\hline
	\end{tabular}
	\caption{Examples of input text, emotions, and generated explanations}
	\label{tab:examples}
\end{table*}
\section{Error Analysis}
\label{errorana}
\subsection{Misclassification of Anger}
Anger is often misclassified due to subtle emotional cues or when it overlaps with related emotions like frustration or anxiety. For example:

\begin{itemize} \item \textbf{Text:} "Man, I can't believe it." \textbf{Explanation:} "The speaker expresses surprise or frustration." \textbf{Predicted:} Anger = 0, Actual = 1. \item \textbf{Text:} "I could not summon up the courage to get up." \textbf{Explanation:} "The speaker conveys vulnerability or exhaustion." \textbf{Predicted:} Anger = 0, Actual = 1. \end{itemize}

These examples indicate that anger is misclassified when the emotional reaction is subtle or related to emotions like frustration or exhaustion, which may not have the overt aggression typically associated with anger.

Additionally, anger is sometimes misclassified due to physical or emotional intensity, which the model may confuse with anxiety or frustration. For example:

\begin{itemize} \item \textbf{Text:} "I felt fire in my stomach." \textbf{Explanation:} "The speaker describes a strong emotional or physical reaction." \textbf{Predicted:} Anger = 0, Actual = 1. \item \textbf{Text:} "There was no stopping the relentless torrent." \textbf{Explanation:} "The speaker describes an intense, unstoppable force." \textbf{Predicted:} Anger = 0, Actual = 1. \end{itemize}

These misclassifications suggest that the system struggles to interpret emotional intensity related to anger, and may categorize it as anxiety or frustration instead.

Lastly, anger is sometimes misclassified as fear or sadness, especially when the emotional cue is indirect or combined with vulnerability:

\begin{itemize} \item \textbf{Text:} "The weekend didn't live up to my storm standards." \textbf{Explanation:} "The speaker expresses disappointment and frustration." \textbf{Predicted:} Anger = 0, Actual = 1. \item \textbf{Text:} "She was growling, barking, snarling, foaming." \textbf{Explanation:} "The speaker describes an intense emotional state, possibly fear or anger." \textbf{Predicted:} Anger = 1, Actual = 0. \end{itemize}

In summary, anger is misclassified due to the subtlety of its expression or its overlap with other emotions such as frustration or anxiety. Additionally, emotional intensity or indirect cues, especially when mixed with vulnerability, can confuse the model. Future improvements should focus on enhancing the model's ability to differentiate between anger and these overlapping emotional states, and better handle the more subtle or complex expressions of anger.
\subsection{Misclassification of Surprise}

Surprise is often misclassified due to subtle or ambiguous emotional cues in the text. For example:

\begin{itemize} \item \textbf{Text:} "The lock was a dial-lock." \textbf{Explanation:} "The speaker describes a specific detail, focusing on the nature of the lock." \textbf{Predicted:} Surprise = 1, Actual = 0. \item \textbf{Text:} "I immediately started getting nervous and panic intensified." \textbf{Explanation:} "The speaker describes anxiety, which may be confused with surprise." \textbf{Predicted:} Surprise = 1, Actual = 0. \end{itemize}

These examples show that Surprise is sometimes misclassified as confusion or anxiety, especially when the emotional reaction is subtle or combined with other emotions.

Additionally, Surprise is occasionally misclassified as fear or anger, particularly when unexpected events are associated with discomfort or frustration:

\begin{itemize} \item \textbf{Text:} "She was growling, barking, snarling, foaming." \textbf{Explanation:} "The speaker describes an intense emotional state, possibly fear or anger." \textbf{Predicted:} Surprise = 1, Actual = 0. \item \textbf{Text:} "I almost got my hands on the door handle, when..." \textbf{Explanation:} "The speaker describes a moment of frustration or missed opportunity." \textbf{Predicted:} Surprise = 1, Actual = 0. \end{itemize}

These misclassifications suggest that when surprise is combined with aggression, frustration, or physical tension, the system may confuse it with fear or anger.

Finally, Surprise is misclassified when there is a lack of clear emotional cues, particularly when surprise is related to unexpected information:

\begin{itemize} \item \textbf{Text:} "My great-grandad was a full-blood Cherokee." \textbf{Explanation:} "The speaker introduces their ancestry with pride and a sense of revelation." \textbf{Predicted:} Surprise = 0, Actual = 1. \end{itemize}

In summary, Surprise is misclassified due to subtle emotional cues, especially when it overlaps with other emotions like fear or anger, or when it is expressed in less overt ways. To improve the model, future work should focus on enhancing its sensitivity to these subtle cues and improving its ability to differentiate Surprise from overlapping emotions.

\end{document}